\documentclass{article}

\PassOptionsToPackage{compress, numbers, sort}{natbib}

\usepackage[final]{tackling_climate_workshop_style}

\usepackage[utf8]{inputenc}  
\usepackage[T1]{fontenc}     
\usepackage{hyperref}        
\usepackage{url}             
\usepackage{booktabs}        
\usepackage{amsfonts}        
\usepackage{nicefrac}        
\usepackage{microtype}       
\usepackage{xcolor}          

\usepackage{graphicx}
\usepackage{authblk}         
\usepackage{booktabs}
\usepackage{tabu}

\newcolumntype{C}[1]{>{\centering\arraybackslash}p{#1}}

\title{MS-nowcasting: Operational Precipitation Nowcasting with Convolutional LSTMs at Microsoft Weather}
\author[$\dagger$]{Sylwester Klocek}
\author[$\dagger$]{Haiyu Dong}
\author[$\mathparagraph$]{Matthew Dixon}
\author[$\dagger$]{Panashe Kanengoni}
\author[$\dagger$]{Najeeb Kazmi}
\author[$\dagger$,*]{Pete Luferenko}
\author[$\dagger$]{Zhongjian Lv}
\author[$\mathparagraph$]{Shikhar Sharma}
\author[$\dagger$]{Jonathan Weyn}
\author[$\dagger$]{Siqi Xiang}
\affil[$\dagger$]{Microsoft Corporation}
\affil[$\mathparagraph$]{Microsoft Turing}
\affil[*]{\texttt{Pete.Luferenko@microsoft.com}}

\begin{document}

\maketitle

\begin{abstract}
  We present the encoder-forecaster convolutional long short-term memory (LSTM) deep-learning model that powers Microsoft Weather's operational precipitation nowcasting product. This model takes as input a sequence of weather radar mosaics and deterministically predicts future radar reflectivity at lead times up to 6 hours. By stacking a large input receptive field along the feature dimension and conditioning the model's forecaster with predictions from the physics-based High Resolution Rapid Refresh (HRRR) model, we are able to outperform optical flow and HRRR baselines by 20-25\% on multiple metrics averaged over all lead times.
\end{abstract}

\section{Motivation}

Accurate short-term forecasts (“nowcasts”) of precipitation are extremely important for many aspects of people’s daily lives, as evidenced by the popularity of services such as Dark Sky. Nowcasting can also identify extreme weather events such as flooding and inform mitigation strategies for these events. With such extreme events predicted to increase in frequency and intensity in a warmer future climate \cite{climate-flooding}, forecasting and early warning will become even more important. 

Traditional forecasting methods based on numerical weather prediction (NWP) models, which compute solutions to differential equations governing the physics of the atmosphere, have a few disadvantages when used for nowcasting. Models such as HRRR \cite{hrrr} are computationally expensive to run at high spatial resolution, cannot predict very short-term precipitation very well due to the required spin-up time of the internal physics, and have limited predictive skill beyond a few hours \cite{durran2016}. Methods for predicting short-term precipitation directly from observed weather radar, including extrapolation methods like optical flow and machine learning methods, have emerged as the state-of-the-art for forecasting precipitation within a few hours. \cite{convlstm} introduced the convolutional long short-term memory (ConvLSTM) architecture with good performance for radar-based precipitation nowcasting. Since then, numerous deep learning architectures and benchmark datasets have been used in a research capacity to tackle this problem \cite[e.g.][]{encoder-decoder-convlstm,lebedev2019precipitation,ayzel2020rainnet,chen2020deep,veillette2020sevir,DGMR}, including Google Research's MetNet \cite{metnet}, which compares favorably against the physics-based HRRR.

Despite the promising performance of deep-learning-based nowcasting, few providers have implemented such systems for operations. We present MS-nowcasting, a ConvLSTM-based model designed for operational efficiency that now powers Microsoft Weather's nowcasting maps and notifications in the US and Europe. Our model has several notable improvements over prior work in precipitation nowcasting, including a novel method to ingest a large spatial input receptive field, and leveraging physics-based HRRR predictions to condition the model's forecaster component. Our design strategies allow MS-nowcasting to run on as little hardware as a single graphics processing unit (GPU). Using only 8 GPUs in production, the model latency for forecasts over the entire US is under two minutes from radar data availability. End-to-end, consumers can enjoy high-quality radar forecast maps within about 6 minutes from the time of radar measurements.

\section{Model}
\begin{figure}
  \centering
  \includegraphics[width=\linewidth]{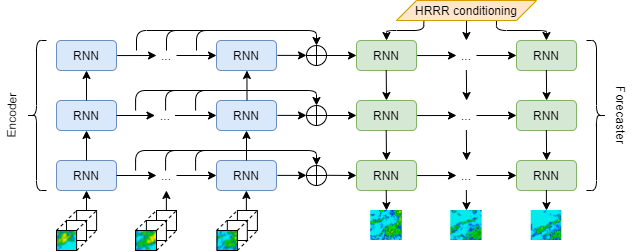}
  \caption{ConvLSTM encoder-forecaster architecture.}
  \label{figure:fig1}
\end{figure}

To target the goal of delivering the best on the market model accuracy and refresh rate while keeping computational resource requirements to a minimum, we needed to make a few key model design choices.
For instance, we needed to avoid expensive operations such as attention mechanism performing $O(n^2)$ operations on tensors. While \citep{metnet} employed the faster approach of axial attention \citep{axial-attention}, it necessitated using 256 Tensor Processing Units (TPUs) running in parallel. 
We decided to leverage existing ConvLSTM architecture first developed in \citep{convlstm} and enhanced in \cite{encoder-decoder-convlstm}, but with some adjustments. 
Our changes to the original model include adding forecaster conditioning based on an NWP model, adding a large viewport of input spatial information in an atypical manner, tuning hyperparameters, weighting of the hidden states, and optimizer changes.

Our model is a three-layer encoder-forecaster network. The input to the model consists of a sequence of $T_i$ images of weather radar reflectivity (a proxy for precipitation intensity) from the Multi-Radar Multi-Sensor dataset \cite[MRMS, ][]{mrms}, with shape $[B, T_i, C, H_i, W_i]$ and the target is a sequence of $T_o$ radar frames, with shape $[B, T_o, 1, H_o, W_o]$. We perform down convolutions followed by leaky ReLU activations before encoder layers and up convolutions followed by leaky ReLU after forecaster layers similar to \citep{encoder-decoder-convlstm}. We keep the original ConvLSTM cell instead of methods like TrajGRU and ConvGRU as we found the latter did not provide significant benefits for our problem. Inside this ConvLSTM cell, we perform group normalization after performing the convolutions. This modified version of the ConvLSTM cell is referred to as RNN in Figure \ref{figure:fig1}. Exact hyperparameters of the model architecture can be found in Table~\ref{table:architecture} in Appendix~\ref{appdxArch}.

Nowcasting models and benchmarks often rely on using the same spatial viewport across the entire sequence of both input and predicted target frames. However, over longer prediction sequences and particularly in rapidly changing weather events, additional spatial context  in the inputs is necessary to provide information about moving precipitation to the model. MetNet \citep{metnet} is one example which uses an input viewport larger than the target; specifically, the authors use an input grid of $1024\times 1024$~km to predict a $64\times 64$~km target square, where 1~km corresponds to one image pixel. In order to fit such a large input array into their model, they perform downsampling and cropping. In contrast, we transform our input viewport of $1280\times 1280$~km  by dividing the tensor of shape $[1280, 1280]$ into 5 equal segments along both the height and width dimensions and stacking them in the channel dimension to obtain a tensor of shape $[25, 256, 256]$. This large viewport is subsequently referred to as LV. Hence, the input neighbourhood is treated as features by our model. The model's predictions are made for the center $256\times 256$~km square. By extensive search of parameters we found that operating on these spatial dimensions is a good trade-off between speed and forecast quality.

Rather than relying solely on radar data, we also condition our MS-nowcasting model by adding another input to the forecaster cells: the reflectivity forecast product from the HRRR model. By using physics of the atmosphere, HRRR is able to predict formation and dissipation of precipitation within the target frames, not just the evolution of existing precipitation. Our model transforms the input HRRR forecast shape to match the forecaster input shapes. First, we linearly interpolate the tensor in the temporal dimension to the desired number of frames $T_o$. Next, we perform a series of down convolutions followed by leaky ReLU operations matching those of the encoder layers. The result is a tensor of shape $[B, T_o, 1, H_{l3}, W_{l3}]$, which can be concatenated directly to the state in the third layer of the forecaster. 

Unlike the previous ConvLSTM publications \citep{convlstm,encoder-decoder-convlstm}, we use all hidden encoder ConvLSTM states to produce forecaster hidden states. For this purpose, we introduce a vector of trainable weights $w = \langle w_{1}, \ldots, w_{m} \rangle$, and each weight has a length of $m$ corresponding to temporal dimension of MRMS input. The forecaster hidden state $h_l$ with shape $[B, 1, C_l, H_l, W_l]$, where $l$ is the layer (up to 3), is computed as 
\begin{equation}
    h_l = \sum_{i=1}^m w_{i} \cdot h_{B, t_i, C_l, H_l, W_l}
\end{equation}
This operation is represented as the $\bigoplus$ symbol in Figure \ref{figure:fig1}. Using more than one past hidden state of the encoder resulted in slight metrics improvements. The cell state of the last encoder RNN cell remains unchanged.

For training purposes, we use the average of mean absolute error (MAE) and mean squared error (MSE) as the loss function, where each frame is weighted by the B-MAE thresholds borrowed from \cite{encoder-decoder-convlstm}. We use Adam optimizer \cite{adam} with stochastic weight averaging \cite[SWA, ][]{swa} on top - this helped our model converge to wider optima, generalize better, and train more stably through mode connectivity \cite{modeconnectivity}. We set the learning rate to $2\times 10^{-4}$, gradient clipping to $1.0$ and weight decay to $1\times 10^{-4}$. For speed and multi-GPU environment training environment, we use the DeepSpeed framework \citep{deepspeed}.

\section{Experiments}
We compare our model to persistence, a naive baseline whereby weather is forecast to remain unchanged, an optical flow method implemented in the \textit{rainymotion} library described in \cite{rainymotion}, and the radar reflectivity forecast from HRRR. 

We perform four ablation experiments to illustrate the improvements achieved with our adjustments to the existing ConvLSTM \cite{convlstm, encoder-decoder-convlstm}.
\begin{enumerate}
    \item \textbf{MS-nowcasting}: baseline model using a small $256\times 256$~km input viewport only.
    \item \textbf{MS-nowcasting + HRRR}: MS-nowcasting with HRRR input to the forecaster.
    \item \textbf{MS-nowcasting + LV}: MS-nowcasting with LV and no HRRR input to the forecaster.
    \item \textbf{MS-nowcasting + HRRR + LV}: MS-nowcasting with both LV and HRRR input to the forecaster.
\end{enumerate}

\subsection{Data}
We prepared the training data by sampling 83,000 sequences of at least 440 minutes from the MRMS archive for 2018, from which we sample 10 million 440-minute windows. The first 80 minutes of these windows are model inputs and the next 360 minutes are targets. In order to reduce compute costs in both training and inference, we use 20 input frames at a 4-minute resolution and 45 target frames at an 8-minute resolution. For the test and validation sets, we drew 1500 and 500 windows, respectively, from the MRMS archive for 2019.

\subsection{Results}

\begin{table}
	\caption{ Metrics averaged over lead times 0--2~hours}
    \label{table:ablation2H}
	\centering
	\begin{tabu}{@{}X[l,1.5,b]*4{X[c,1,b]}@{}}
		\toprule
		Model & MAE & F1 (12 dBZ) & MS-SSIM & PSNR \\ \midrule
		Persistence   & 4.88 & 0.686 & 0.5156 & 21.21 \\
		Optical Flow  & 4.39 & 0.720 & 0.5527  & 21.83 \\
		HRRR          & 5.54 & 0.610 & 0.4634  & 20.01 \\ \cmidrule{2-5}
		MS-nowcasting & 3.64 & 0.788 & 0.6419  & 23.23 \\
		~ + HRRR      & 3.55 & 0.780  & 0.6510  & 23.29 \\
		~ + LV        & 3.28 & 0.808 &  0.6678  & 23.86 \\
		~ + HRRR + LV & \textbf{3.23} & \textbf{0.813} &  \textbf{0.6721}  & \textbf{24.03} \\ \bottomrule
   \end{tabu}
\end{table}

\begin{figure}
  \centering
  \includegraphics[width=\linewidth]{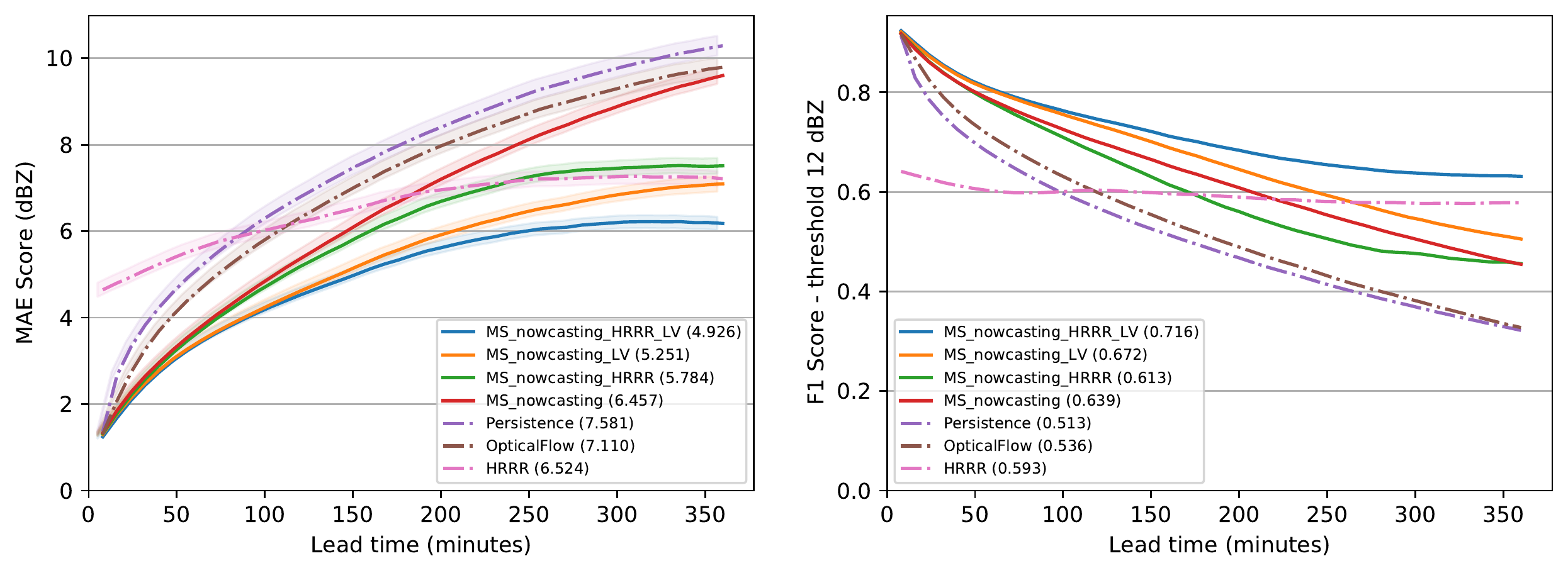}
  \caption{MAE and F1 metrics over lead times for the ablation study. Numbers in parentheses in the plot legends are the average metric value over all lead times.}
  \label{figure:fig2}
\end{figure}

We evaluate the mean absolute error (MAE) and F1 score at the threshold of 12 dBZ (corresponding to precipitation rates of approximately 0.2 mm/h) of the predictions against ground truth. We also evaluate the Multiscale Structural Similarity Index Measure (MS-SSIM) and Peak Signal-to-Noise Ratio (PSNR) to measure image quality relative to ground truth. Table \ref{table:ablation2H} shows that our model (MS-nowcasting + HRRR + LV) substantially outperforms all others on all metrics in the first two hours, while even MS-nowcasting is better than all three baselines in the same period. 

Figure \ref{figure:fig2} shows the metrics by lead time for the ablation study. While MS-nowcasting starts off better than HRRR, its performance becomes worse at longer lead times. Adding HRRR input improves performance on MAE, but worsens it on F1. This is explained by the model bias presented in Figure \ref{figure:fig3} in Appendix~\ref{appdxMetrics}. MS-nowcasting has a high overprediction bias, especially at long lead times. Conversely, adding HRRR leads to a high underprediction bias, which causes the model to miss the 12 dBZ threshold more frequently. On the other hand, adding LV alone gives more gains than adding HRRR alone as that model has less of an underprediction bias.

The final model with both LV and HRRR outperforms all the rest at all lead times, demonstrating that both larger spatial context and formation and dissipation of precipitation provided by HRRR are important in predicting longer sequences of precipitation. Nevertheless, the choice of MAE+MSE loss for the model results in predictions that appear overly smooth (see Figs.~\ref{sample1} and \ref{sample2} in the Appendix). Recent studies have shown that generative adversarial networks are a potential way of addressing this deficiency \cite{DGMR}.

\section{Conclusion}
In this paper, we presented MS-nowcasting, an encoder-forecaster long short-term memory (LSTM) deep-learning model for precipitation nowcasting. MS-nowcasting improves upon the existing ConvLSTM \cite{convlstm, encoder-decoder-convlstm} architecture and outperforms the operational HRRR NWP model for lead times of up to 6 hours. Our approach allows for efficient and accurate rapid-update precipitation nowcasting. The model is operational and can be potentially inform forecasts for early warning of extreme weather events.

\begin{ack}
The authors thank the many engineers at Microsoft who helped make the production model a reality.
\end{ack}

{
    \small
    \bibliographystyle{unsrt}
    \bibliography{precipitation-nowcasting}
}

\appendix
\newpage
\section{\label{appdxMetrics}Appendix: Additional metrics}

\begin{table}[!h]
	\caption{Metrics averaged over all lead times 0--6~hours}
    \label{table:ablation6h}
	\centering
	\begin{tabu}{@{}X[l,1.5,b]*4{X[c,1,b]}@{}}
		\toprule
		Model & MAE & F1 (12 dBZ) & MS-SSIM & PSNR \\ \midrule
		Persistence   & 7.58 & 0.513 & 0.3626 & 16.98 \\
		Optical Flow  & 7.11 & 0.536 & 0.3780 & 17,38 \\
		HRRR          & 6.52 & 0.593 & 0.3905  & 17.72 \\ \cmidrule{2-5}
		MS-nowcasting & 6.46 & 0.639 & 0.4788  & 18.78 \\
		~ + HRRR      & 5.78 & 0.672  & 0.4952  & 19.10 \\
		~ + LV        & 5.25 & 0.613 &  0.5137  & 19.68 \\
		~ + HRRR + LV & \textbf{4.92} & \textbf{0.716} &  \textbf{0.5291}  & \textbf{20.16} \\ \bottomrule
   \end{tabu}
\end{table}

\begin{figure}[!h]
  \centering
  \includegraphics[width=\linewidth]{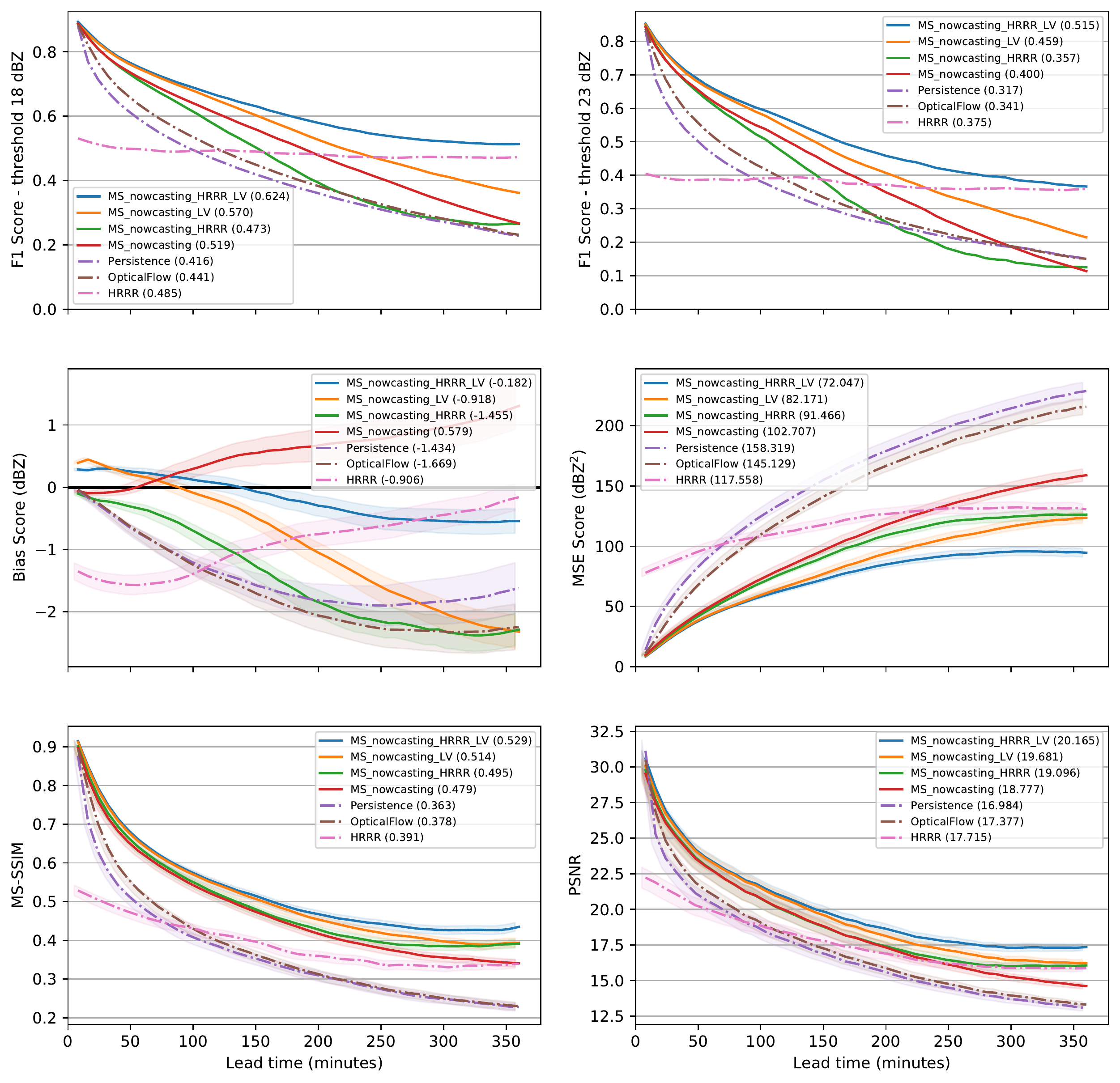}
  \caption{As in Fig.~\ref{figure:fig2}, but including more metrics (as labeled). F1 thresholds of 18 and 23 dBZ correspond to precipitation rates of roughly 0.5 and 1.0 mm/h respectively.}
  \label{figure:fig3}
\end{figure}

\newpage
\section{\label{appdxArch}Appendix: Architecture}

\begin{table}[h]
	\caption{Model architecture}
    \label{table:architecture}
	\centering
	\small
	\begin{tabu}{p{0.42\textwidth}C{0.06\textwidth}C{0.2\textwidth}C{0.06\textwidth}C{0.07\textwidth}}
		\toprule
		Name & Kernel size & Size & Stride & Padding  \\ \midrule
		hidden state weights   & - & [20] & - & - \\ \cmidrule{2-5}
		L0-encoder-downsconv-weight & 6x6 & [16, 25, 6, 6] & 3  & 0 \\
		L0-encoder-downsconv-bias & - & 16 & -  & - \\ 
		L0-encoder convlstmcell $i_g, f_g, c_g, o_g$  & 3x3 & [256, 80, 3, 3] & 1 & 1 \\
		L0-encoder convlstmcell biases  & - & [256]  & -  & - \\
		L0-encoder groupnorm weight & - & [256] &  -  & - \\
		L0-encoder groupnorm bias & - & [256] &  -  & - \\ \cmidrule{2-5}
		L1-encoder-downsconv-weight & 5x5 & [192, 64, 5, 5] & 3  & 1 \\
		L1-encoder-downsconv-bias & - & [192] & -  & - \\ 
		L1-encoder convlstmcell $i_g, f_g, c_g, o_g$  & 3x3 & [768, 384, 3, 3] & 1 & 1 \\
		L1-encoder convlstmcell biases  & - & [768]  & -  & - \\
		L1-encoder groupnorm weight & - & [768] &  -  & - \\
		L1-encoder groupnorm bias & - & [768] &  -  & - \\ \cmidrule{2-5}
		L2-encoder-downsconv-weight & 3x3 & [192, 192, 3, 3] & 2 & 1 \\
		L2-encoder-downsconv-bias & - & [192] & -  & - \\ 
		L2-encoder convlstmcell $i_g, f_g, c_g, o_g$  & 3x3 & [768, 384, 3, 3] & 1 & 1 \\
		L2-encoder convlstmcell biases  & - & [768]  & -  & - \\
		L2-encoder groupnorm weight & - & [768] &  -  & - \\
		L2-encoder groupnorm bias & - & [768] &  -  & - \\ \cmidrule{2-5}
		HRRR conditioning downconv-0 weight  & 6x6 & [16, 1, 6, 6] & 3  & 0 \\
		HRRR conditioning downconv-0 bias  & - & [16] & -  & - \\ 
		HRRR conditioning downconv-1 weight  & 6x6 & [192, 16, 5, 5] & 3  & 1 \\
		HRRR conditioning downconv-1 bias  & - & [192] & -  & - \\ 
		HRRR conditioning downconv-2 weight  & 3x3 & [192, 192, 3, 3] & 2  & 1 \\
		HRRR conditioning downconv-2 bias  & - & [192] & -  & - \\ 
		 \cmidrule{2-5} 
		L2-forecaster-convlstmcell $i_g, f_g, c_g, o_g$ & 3x3 & [768, 384, 3, 3] & 1 & 1  \\
		L2-forecaster-convlstmcell bias & - & [768] & -  & - \\ 
		L2-forecaster groupnorm weight  & - & [768] & - & - \\
		L2-forecaster groupnorm biases  & - & [768]  & -  & - \\
		L2-forecaster upconv weight & 4x4 & [192, 192, 4, 4] &  2  & 1 \\
		L2-forecaster upconv bias & - & [192] &  -  & - \\ \cmidrule{2-5}
		L1-forecaster-convlstmcell $i_g, f_g, c_g, o_g$ & 3x3 & [768, 384, 3, 3] & 1 & 1  \\
		L1-forecaster-convlstmcell bias & - & [768] & -  & - \\ 
		L1-forecaster groupnorm weight  & - & [768] & - & - \\
		L1-forecaster groupnorm biases  & - & [768]  & -  & - \\
		L1-forecaster upconv weight & 5x5 & [192, 64, 5, 5] &  3  & 1 \\
		L1-forecaster upconv bias & - & [64] &  -  & - \\ \cmidrule{2-5}
		L0-forecaster-convlstmcell $i_g, f_g, c_g, o_g$ & 3x3 & [256, 128, 3, 3] & 1 & 1  \\
		L0-forecaster-convlstmcell bias & - & [256] & -  & - \\ 
		L0-forecaster groupnorm weight  & - & [256] & - & - \\
		L0-forecaster groupnorm biases  & - & [256]  & -  & - \\
		L0-forecaster upconv weight & 7x7 & [64, 16, 7, 7] &  3  & 0 \\
		L0-forecaster upconv bias & - & [16] &  -  & - \\ \cmidrule{2-5}
        final-conv.0.weight & 3x3 & [16, 16, 3, 3] & 1  & 1 \\
        final-conv.0.bias  & - & [16]  & -  & - \\
        final-conv.2.weight & 1x1 & [1, 16, 1, 1] & 1 & 0 \\
        final-conv.2.bias & - & [1] &  -  & - \\ \bottomrule
   \end{tabu}
\end{table}

\newpage
\section{\label{AppdxSamples}Appendix: Sample predictions}

\begin{figure}[!h]
  \includegraphics[width=\linewidth]{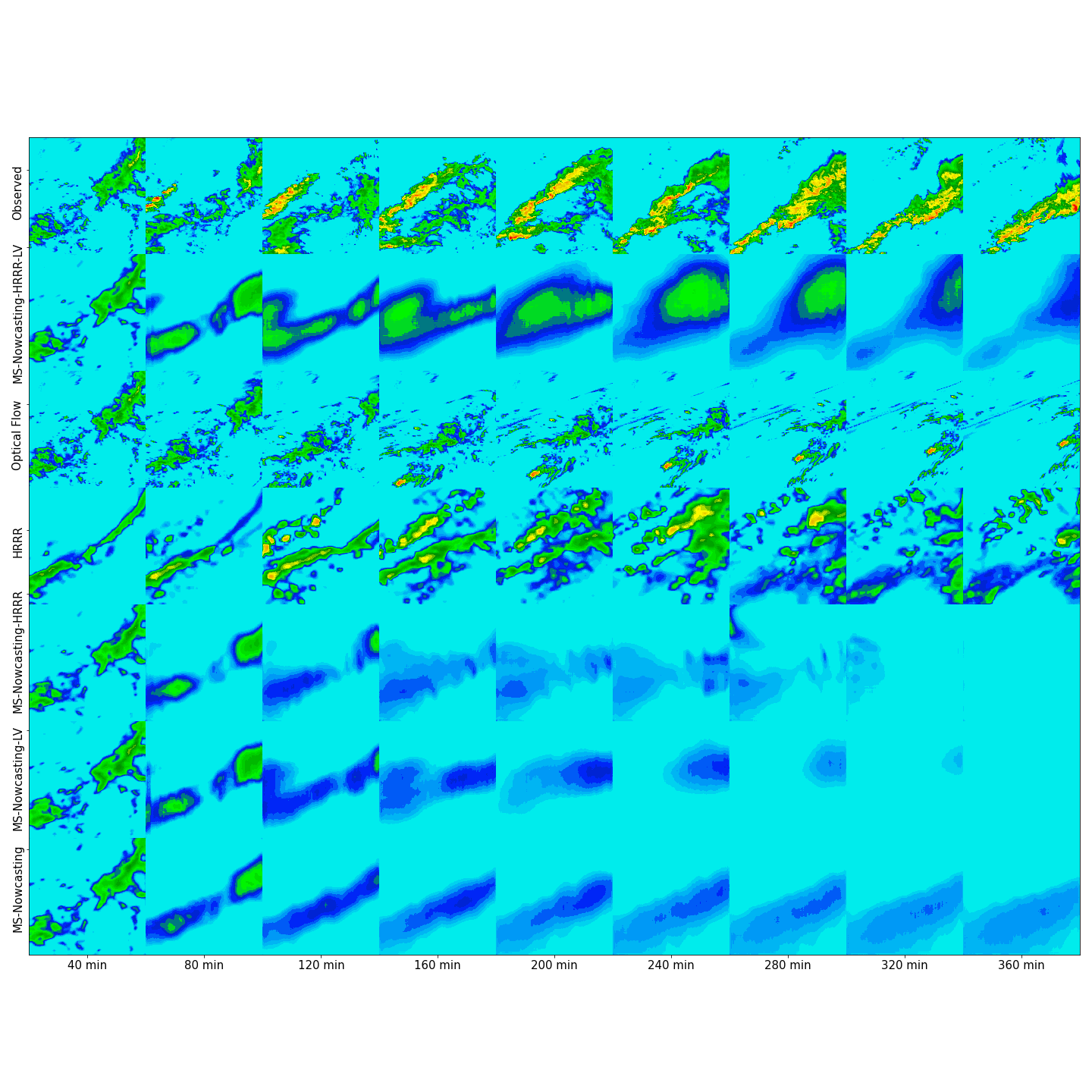}
  \caption{An example sequence of observed true radar reflectivity (top row) and predicted radar from our models and the optical flow and HRRR baselines, as labeled.}
  \label{sample1}
\end{figure}

\begin{figure}[!h]
  \includegraphics[width=\linewidth]{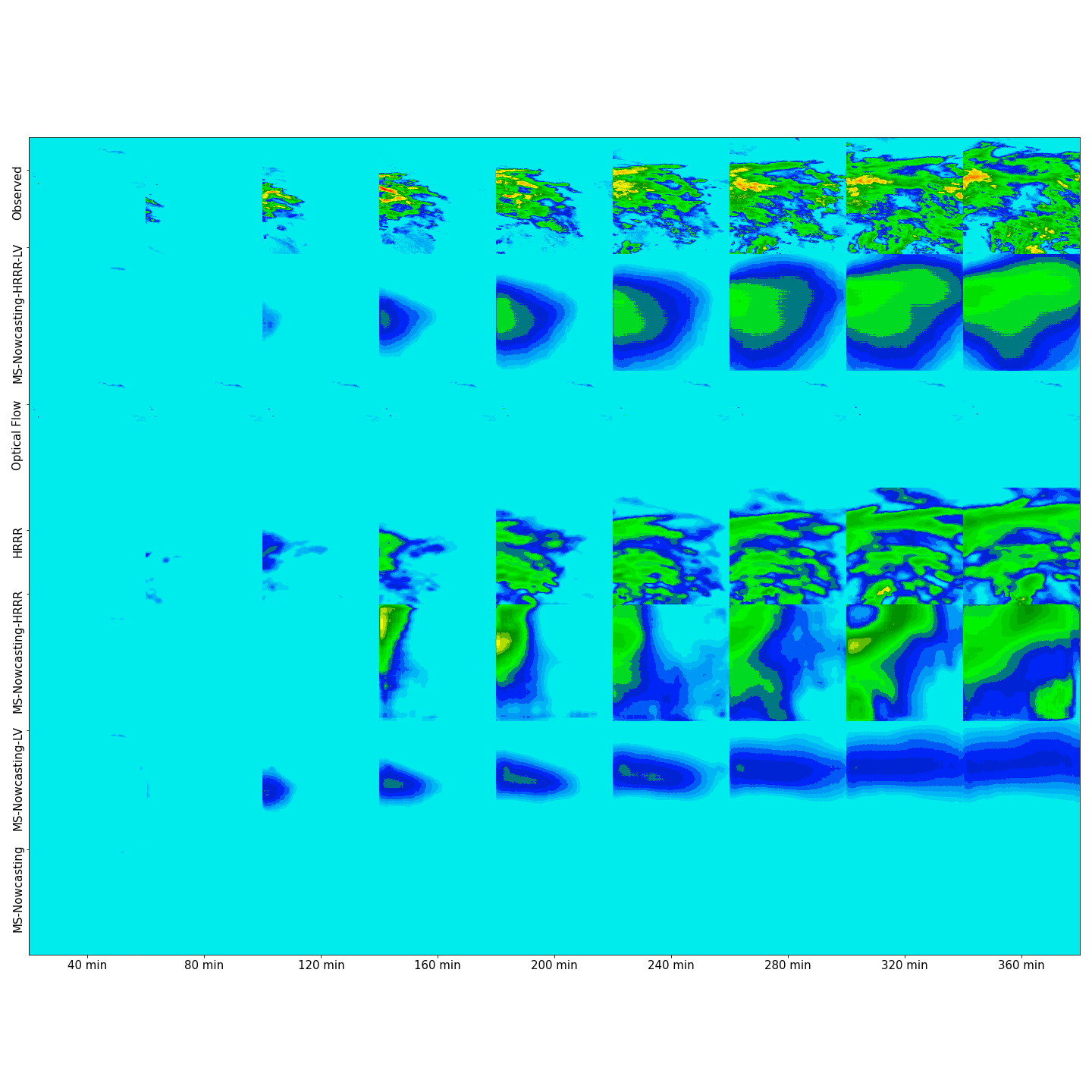}
  \caption{As in Fig.~\ref{sample1}, but a different example sequence.}
  \label{sample2}
\end{figure}

Figures~\ref{sample1} and \ref{sample2} show two example sequences of predicted and observed true radar reflectivity from the test set. In both cases, the precipitation evolves substantially over the 6-hour forecast period. In Fig.~\ref{sample1}, a new line of thunderstorms develops behind (to the northwest of) the original band of precipitation and intensifies. The HRRR model generally captures this evolution but produces too scattered and weak thunderstorms. Our best MS-nowcasting + HRRR + LV captures the general behavior of the precipitation pattern but its forecasts are too smooth and weak. Meanwhile, the other MS-nowcasting models without the LV and/or HRRR conditioning greatly dissipate the rain, while the raw HRRR predicts precipitation in incorrect locations.

In Fig.~\ref{sample2}, rain is progressing from west to east into the target region while also intensifying. The optical flow baseline, despite containing the full 1280$\times$1280 input, fails to capture any precipitation, as does the MS-nowcasting model with no large viewport. Meanwhile the MS-nowcasting + HRRR + LV model accurately picks up on the large area of precipitation, although it loses most of the detailed structure.


\begin{figure}[!h]
  \includegraphics[width=\linewidth]{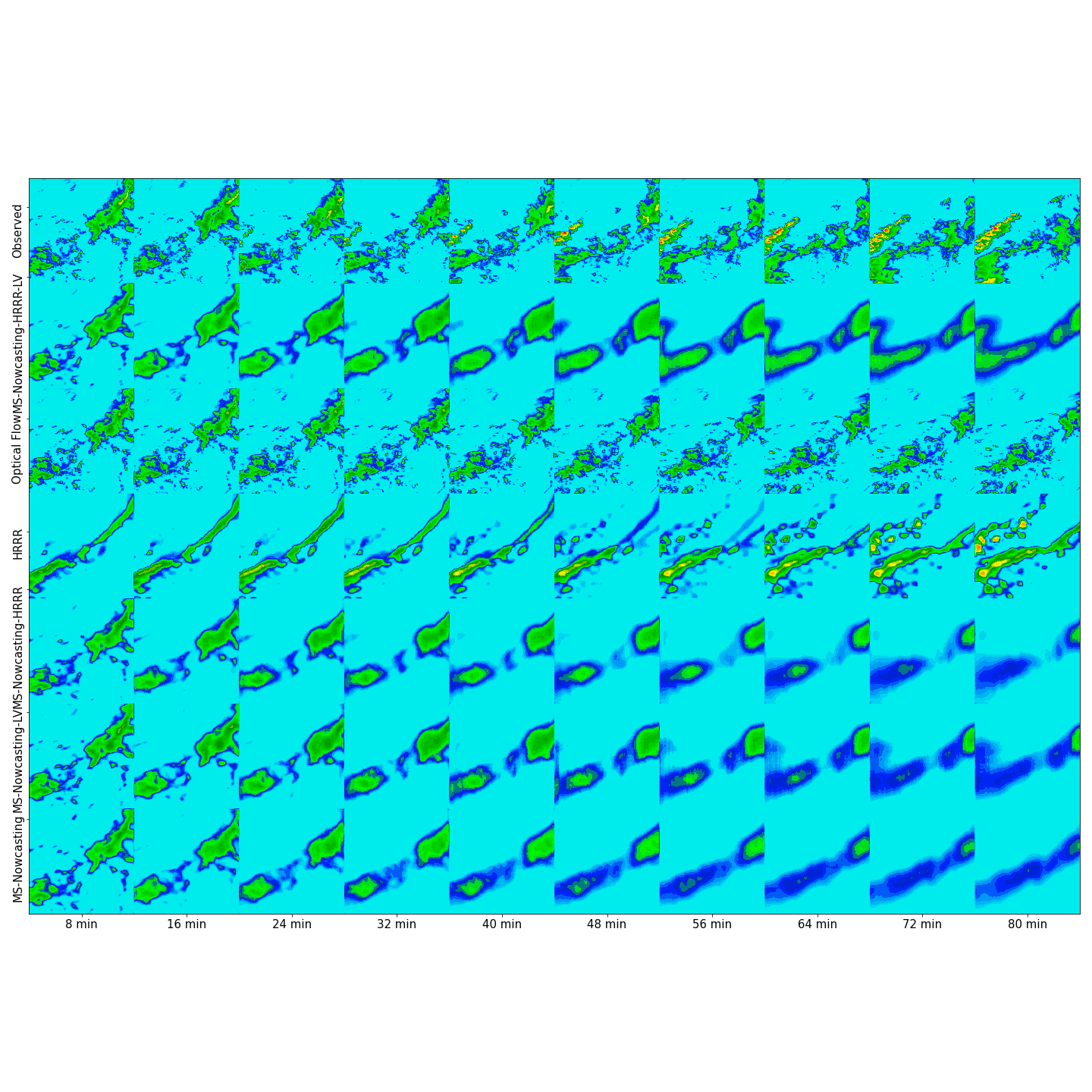}
  \caption{An in Fig .~\ref{sample1}, but predictions for 80 mins into the future with 8 min interval.}
  \label{sample3}
\end{figure}

\end{document}